\documentclass{article}

\usepackage{microtype}
\usepackage{graphicx}
\usepackage{booktabs} 

\usepackage{hyperref}

\newcommand{\ourmethod}{\dtsne}
\newcommand{\oururl}{\url{http://eda.rg.cispa.io/dtsne/}}

\usepackage{latexsym}
\usepackage[ruled, vlined, nofillcomment, linesnumbered]{algorithm2e}
\usepackage[notheorems]{eda}
\usepackage{subcaption}
\usepackage{array,multirow}

\usepackage{savesym}
\usepackage{cancel}
\usepackage{textgreek}

\graphicspath{ {./figs/} }

\usepackage{palettes}
\usepackage{pgfplots}
\usepgfplotslibrary{external}
\usetikzlibrary{patterns,positioning}

\newcommand{\norm}[1]{\left\lVert#1\right\rVert}

\newcommand{\tsne}{\textsc{tSNE}\xspace}
\newcommand{\sne}{\textsc{SNE}\xspace}
\newcommand{\dtsne}{\textsc{dtSNE}\xspace}
\newcommand{\umap}{\textsc{UMAP}\xspace}
\newcommand{\lle}{\textsc{LLE}\xspace}
\newcommand{\largevis}{\textsc{LargeVis}\xspace}
\newcommand{\ncvis}{\textsc{NCVis}\xspace}

\makeatletter
\newcommand\incircbin
{%
  \mathpalette\@incircbin
}
\newcommand\@incircbin[2]
{%
  \mathbin%
  {%
    \ooalign{\hidewidth$#1#2$\hidewidth\crcr$#1\bigcirc$}%
  }%
}

\makeatother

\pgfkeys{/pgfplots/tuftelike/.style={
  semithick,
  tick style={major tick length=4pt,semithick,black},
  separate axis lines,
  axis x line*=bottom,
  axis x line shift=10pt,
  xlabel shift=5pt,
  axis y line*=left,
  axis y line shift=10pt,
  ylabel shift=5pt}}

\pgfplotsset{compat=newest}

\usepackage[accepted]{icml2023}

\usepackage{amsmath}
\usepackage{amssymb}
\usepackage{mathtools}
\usepackage{amsthm}

\usepackage[capitalize,noabbrev]{cleveref}

\theoremstyle{plain}

\theoremstyle{definition}

\theoremstyle{remark}

\usepackage[disable,textsize=tiny]{todonotes}

\def\useprebuildfigures{} 

\icmltitlerunning{Preserving local densities in low-dimensional embeddings}

\begin{document}

\twocolumn[
\icmltitle{Preserving local densities in low-dimensional embeddings}



\icmlsetsymbol{equal}{*}

\begin{icmlauthorlist}
\icmlauthor{Jonas Fischer}{yyy}
\icmlauthor{Rebekka Burkholz}{sch}
\icmlauthor{Jilles Vreeken}{sch}
\end{icmlauthorlist}

\icmlaffiliation{yyy}{Department of Biostatistics, Harvard T.H. Chan School of Public Health, Boston MA, USA}
\icmlaffiliation{sch}{CISPA Helmholtz Center for Information Security, Saarbrücken, Germany}

\icmlcorrespondingauthor{Jonas Fischer}{jfischer@hsph.harvard.edu}

\icmlkeywords{low-dimensional embeddings, manifold learning, visualization, tSNE, UMAP}

\vskip 0.3in
]



\printAffiliationsAndNotice{}  

\begin{abstract}
Low-dimensional embeddings and visualizations are an indispensable tool for analysis of high-dimensional data.
State-of-the-art methods, such as \tsne and \umap, excel in unveiling local structures hidden in high-dimensional data and are therefore routinely applied in standard analysis pipelines in biology.
We show, however, that these methods fail to reconstruct local properties, such as relative differences in densities (Fig.~\ref{fig:motivation_density}) and that apparent differences in cluster size can arise from computational artifact caused by differing sample sizes (Fig.~\ref{fig:motivation_samples}).
Providing a theoretical analysis of this issue, we then suggest \ourmethod, which approximately conserves local densities.
In an extensive study on synthetic benchmark and real world data comparing against five state-of-the-art methods, we empirically show that \ourmethod provides similar global reconstruction, but yields much more accurate depictions of local distances and relative densities.
\end{abstract}

\section{Introduction}

\input{figs/gaussian2d_densities}
\input{figs/gaussian2d_samples}

Low-dimensional embeddings are an essential tool of data analysis allowing exploration of the structure and relationships encoded in the data.
Given the high-dimensional datasets that are gathered on a daily basis, such low-dimensional embeddings have been shown to be especially fruitful in aiding experts to identify general trends, clusters and inter-cluster relationships, as well as extreme-valued samples and outliers.
In natural sciences, such as genomics, they are routinely applied as a first step in data exploration, in core machine learning they are frequently used as a tool for understanding neural embeddings, such as given by word or sentence encoders.
The property of most high-dimensional data that allows for such a reduction of dimensions is that samples live in a lower dimensional subspace or manifold, which low-dimensional embeddings aim to approximate.

While a classical  projection onto the first two principal components  can be insightful and a valuable first step of analysis, the interesting regularities are often non-linear and are, hence, not approximated well by principal components.
Different solutions have been suggested for this problem, with most successful advances using a similarity measure for the high-dimensional data to be reconstructed in the low-dimensional embedding.
Moreover, it is inherently hard to correctly reconstruct relationships at all distance scales.
Instead, state-of-the-art methods focus on reconstructing local structures correctly, in exchange allowing to distort long-range distances, as this arguably preserves the more interesting regularities in the data.
The most widely used are \tsne~\cite{maaten2008visualizing} and \umap~\cite{mcinnes2018umap}, as well as their recent adapatations \largevis~\cite{tang:16:largevis} and \ncvis~\cite{artamenkov:20:ncvis}, which are by now an indispensable part in standard data analysis pipelines in, e.g., biology.

While these methods yield arguably good overall reconstructions and visualizations of the data, their embeddings share an often neglected problem.
Most experts are well aware that inter-cluster distances are distorted and argue about those with care. 
In contrast, individual clusters are (at least relative to each other) often assumed to be faithfully reconstructed locally.

This, however, turns out not to be true: Cluster sizes and densities in the embedding do not model variances respectively densities of the high-dimensional data.
Two clusters with the same number of points, as diverse as one stretched over vasts amount of space and the other one extremely compact, both appear equally sized for current embedding techniques, which we visualize in a simple example in Fig.~\ref{fig:motivation_density}.
In experiments on real world data, these methods often yield embeddings with different cluster sizes, which, however, turn out to be not due to differently sized clusters in high dimensions, but to be different numbers of samples per cluster artificially bloating clusters in the embedding (see Fig.~\ref{fig:motivation_samples}).

These relative differences in local densities, which state-of-the-art methods fail to model correctly, however, could provide crucial information about the data at hand, such as a cell types, which are evident as clusters, being more heterogeneous than others in a biological dataset.
Here, we provide a theoretical argument why current methods fail to reconstruct relative densities and suggest a domain-independent and entirely unsupervised approach to properly account for relative local densities.
In extensive experiments on synthetic and real world data, we show that our solution  better recovers the relative differences of cluster densities hidden in the high dimensional data, while yielding overall performance that is on par with state-of-the-art approaches in terms of overall reconstruction quality.

In summary, our main contributions are
\begin{itemize}
    \item a theoretical analysis of why current methods do not reflect local densities correctly,
    \item an unsupervised embedding approach preserving relative densities (\dtsne),
    \item a theoretical analysis of \dtsne proving its ability to preserve densities for local neighbourhoods, and
    \item an extensive evaluation on synthetic and real-world data against six state-of-the-art methods.
\end{itemize}

\section{Related Work}\label{sec:related}

Embeddings of high dimensional data into a low dimensional space, in particular to 2 or 3 dimensions, have in recent years become an essential tool of unsupervised analysis of high dimensional data in modern science.
Classical methods of principal component analysis~\cite{pearson1901liii}, multidimensional scaling~\citep{torgerson1952multidimensional}, laplacian eigenmaps~\cite{belkin:01:laplacian}, and self organizing maps~\citep{kohonen1982self} focus on keeping all, and in particular the large distances intact.
As high dimensional data typically lies on a manifold~\citep{silva2003global}, resembling euclidean space only locally, research attention shifted on modeling geodesic distances~\citep{tenenbaum2000global}, or focusing only on local distances trough locally linear embeddings (\lle)~\citep{roweis2000nonlinear} or stochastic neighbour embeddings (\sne)~\citep{hinton2003stochastic}.

The current state-of-the-art in low-dimensional embeddings for visualizations t-distributed SNE (\tsne) by \citet{maaten2008visualizing} and Uniform Manifold Approximation (\umap) by \citet{mcinnes2018umap} also focus on modeling local distances correctly, allowing to distort long-range distances.
They successfully reveal intrinsic structures of high dimensional data and, hence, have been adapted as standard exploration and pre-processing tools in, e.g., genomics~\citep{becht2019dimensionality, kobak2019art} and embeddings through natural language processing~\citep{coenen2019visualizing}. As they usually yield highly similar embeddings when properly initialized~\citep{kobak:21:tsneumapsame}, it is a matter of taste which one to use, especially after recent algorithmic improvements~\citep{linderman2019fast} also resulted in similar runtimes.
Recent theoretical works confirmed that the clustering revealed by \tsne is provably correct under simple assumptions about the data~\cite{arora2018tsneanalysis, linderman:19:tsneprovably} and unified the theory behind both \tsne and \umap through the lens of contrastive learning~\cite{damrich:22:unifiedumaptsne}.

Building on the successful application of low dimensional embeddings to real world problems,
\citet{kobak2019heavy} extended \tsne to reveal more pronounced and fine-grained cluster structures. 
To emphasize user-specified structures, several works proposed supervised and interactive approaches~\citep{de2003supervised, alipanahi2011guided, barshan2011supervised}. Similarly, supervised approaches were designed to account for unwanted or known variations -- revealing knowledge beyond what is already known -- based on user interaction~\citep{puolamaki2018interactive} or by removing information given by a prior~\citep{kang2016subjectively, kang:21:cond, heiter:21:confetti}.
To extend low-dimensional embeddings to extremely large-scale datasets in terms of samples, such as given by huge web-crawls of newspapers, recent advances focused on improving the runtime of different embedding techniques~\cite{tang:16:largevis, artamenkov:20:ncvis}.

We are specifically interested in unsupervised low dimensional embeddings and hence compare to the state-of-the-art methods \tsne, \umap, \lle, \largevis, and \ncvis.

\section{Theory}

In this section, we first provide a description of the general problem statement for low dimensional embeddings. We then revisit the formulation of t-distributed stochastic neighbour embedding (\tsne) thereby analyzing its inherent limitations with regard to preserving (relative) local densities. Subsequently, we propose \ourmethod, an adaptation which properly addresses modeling local densities.

\subsection{Low-dimensional embeddings}

Given data, $X=(x_1, x_2,\ldots,x_n)$ of $n$ samples, where $x_i \in \mathbb{R}^{m}$ is usually of large dimension $m$, our aim is to find an embedding $Y=(y_1, y_2,\ldots,y_n)$, $y_i \in \mathbb{R}^{m'}$ where  $m'\in\{2,3\}$, such that the important structure in $X$ is preserved in $Y$.
To model the structure in $Y$, first approaches suggested to preserve the relative pairwise distances, i.e. $\lambda\norm{y_i - y_j}\approx\norm{x_i - x_j}$, over a norm based on euclidean or geodesic distances for some algorithm dependent scaling factor $\lambda$, which could be e.g. based on normalizing $X$.
Preserving all (relative) distances, however, is usually not possible and leads to poor results on complex data~\citep{maaten2008visualizing} as a lower dimensional space can encode less information.
Consider the simple example of 4 clusters living in some higher dimensional space, all equidistant to each other, then modeling all distances between clusters correctly becomes infeasible in 2 dimensions.\!\footnote{This can be immediately seen from Pythagoras' theorem.} With the typically used pairwise Euclidean distance, which emphasize modeling large distances correctly, the clusters get distorted to maintain the inter-cluster distances.

Preserving relative local distances and only crudely approximating global distances offers a remedy, as it shows to preserve the relevant structure of the data despite the loss of information~\citep{maaten2008visualizing, mcinnes2018umap}.
In the example before, this would mean to only approximating the inter-cluster distances, but keeping the important local distances -- the within-cluster distances -- intact.
Next, we revisit \tsne, which is considered state-of-the-art for low-dimensional embeddings.

\subsection{\tsne}

The t-distributed stochastic neighbour embedding models the relationship between points $i$ and $j$ in the high dimensional space $X$ and embedding $Y$ in terms of a similarity metric that emphasizes local structure. In particular, similarity of $i,j$ in $X$ is modeled by a normalized Gaussian centered at $x_i$
\[
 p_{j|i} = \frac{\exp(-\norm{x_i - x_j}_2^2/(2\sigma_i^2))}{\sum_{k\neq i}\exp(-\norm{x_i - x_k}_2^2/(2\sigma_i^2))}\; .
\]
Through the normalization in the denominator it can be interpreted as the conditional probability that $i$ would pick $j$ as its neighbour, if neighbours were picked in proportion to their probability density under a Gaussian centered at $x_i$.
The conditional probabilites over all $j$ given an $i$ yield the probability distribution $P_i$.

The deviation $\sigma_i$ of sample $i$ is fitted to account for the differences in local densities across the data, i.e., for dense regions smaller values of $\sigma_i$ suffice to capture local structure, whereas we need large $\sigma_i$ for sparser regions to capture enough of the local structure.
In practice, \tsne employs a binary search to find the $\sigma_i$  that produces a $P_i$ for fixed perplexity.
The perplexity of a distribution $P_i$ is defined as $\mathit{perplexity}(P_i)=2^{H(P_i)}$, where $H(P_i)=-\sum_j p_{j|i}\log p_{j|i}$ is the Shannon entropy.
In practice, it is a user-set hyperparameter and can be seen as a smooth measure of effective number of neighbours, for which we can solve above equation for $\sigma_i$. For more information, we refer to~\citet{maaten2008visualizing}.
Plugging the $\sigma_i$ into the definition of $p_{j|i}$ above, we obtain a probability distribution for each point
$
P_i=\sum_{j\neq i} p_{j|i} \; .
$

To ease computation, probabilities are symmetrized to obtain one joint probability $P$, where for $n$ samples we get
\[
 p_{ij}= \frac{p_{j|i} + p_{i|j}}{2n}\;.
\]

The corresponding low dimensional probabilities are modeled by a t-distribution with one degree of freedom
\begin{equation}\label{eq:lowdim} 
q_{ij} = \frac{(1+\norm{x_i - x_j}_2^2)^{-1}}{\sum_{k\neq l}(1+\norm{x_k - x_l}_2^2)^{-1}}\; ,
\end{equation}
which -- in contrast to the Gaussian distribution -- allows to model large distances more flexible due to its heavy tails, thus avoiding the crowding problem\footnote{The crowding problem is the phenomenon of assembling all points in the center of the map, due to the accumulation of many small attractive forces as moderate distances are not accurately modelled.} in the embedding space~\citep{maaten2008visualizing}.

With representations of similarities for the given high dimensional data, and the (unknown) embedding, both in terms of probability distributions, we can now optimize for the probability distribution $Q$ to model $P$.
In \tsne, this is done by optimizing the Kullback-Leibler (KL) divergence
\begin{equation}\label{eq:kl}
KL(P ~||~ Q) = \sum_i \sum_j p_{ij} \log\left(\frac{p_{ij}}{q_{ij}}\right)\;. 
\end{equation}
The KL divergence measures the number of additional bits needed to encode $P$ using a code optimal for encoding $Q$ and thus models how well $Q$ approximates $P$.
It is differentiable with respect to $y_i$ and thus allows for optimization via gradient descent approaches.
A common problem of state-of-the-art embedding techniques, including \tsne, is, however, that relative densities in the data $X$, such as differently sized clusters or different point densities within clusters, are not captured in the embedding $Y$.

\subsection{One size does not fit all}

The state-of-the-art methods for low-dimensional embeddings, such as \tsne, \ncvis, \largevis, and \umap, usually model low-dimensional distances locally as we have seen before, using the same representation of those distances for different regions in the space, regardless of local densities. 
As an example, consider \tsne, which maps distances represented as $p_{ij}$, which reflects  neighbourhood density through the learned $\sigma_i, \sigma_j$, to $q_{ij}$, which is scale-less. 
Hence, no matter how far stretched, or how compact a cluster is, it will be assigned the same amount of space in low dimensions.
This is what we see happening in Fig.~\ref{fig:motivation_density}.

Why, however, do we see differently sized clusters in embeddings of \tsne and \umap?
For data of clusters with same density (or variance), but varying number of samples per cluster, we observe that \tsne and \umap embed these clusters as vastly differently sized in the embedding space (see Fig.~\ref{fig:motivation_samples}).
What happens is that in high dimensions when the local variance $\sigma_i$ is very small, the resulting Gaussian $P_i$ is hence very narrow. The more points we have in a cluster, the more likely it is that we have $k$NNs that are very close to $i$ and hence the neighbourhood distribution $P_i$ becomes narrow. Points from the same cluster, but which are further away than the $k$NNs of $i$, hence fall into the tail of $P_i$ and have to be matched with a similar probability mass in $Q$. This also means that they are modeled further away than their counterparts in clusters with fewer points, where the $k$NNs are further apart.

While in these arguments we discussed the formulation of \tsne, the same applies for \ncvis, \largevis, and \umap. \ncvis and \largevis follow a analogous formulation of neighbourhood probabilities as \tsne. Although \umap has solid theoretical foundations in Riemannian geometry and fuzzy simplicial sets, its practical implementation has a one-to-one correspondence to \tsne as discussed by the original authors~\citep[Appendix C]{mcinnes2018umap}.
In particular, the low dimensional distribution also has the same shape for each data point, which is why \umap suffers from the same issues as \tsne.

Next, we provide a theoretical argument that \tsne fails to capture variations of local densities.

\subsection{Reflecting local densities in embeddings}

Theoretical insights into \tsne are rare, as the method 
and the solutions to the related optimization problem do not need to be unique.
Gradient descent, which is usually employed in this context, only seeks for a local minimum that has zero gradient.
To still be able to reason about our method we instead argue about those solutions that best support the intuition behind the design of \tsne, which is to approximate local distances well, and that we would prefer to find with our optimization approach.
Those solutions should fulfill the zero gradient condition by $q_{in} \approx p_{in}$.

Let us consider two sufficiently small distances $\norm{x_i - x_j}$ and $\norm{x_k - x_l}$ so that $\norm{x_i - x_j}^2 < \epsilon$ and $\norm{x_k - x_l}^2 < \epsilon$ for a small $\epsilon > 0$.
All involved points $x_p$ for $p \in \{i,j,k,l\}$ have lower dimensional representations $y_i$, $y_j$, $y_k$, $y_l$ that are obtained from an embedding method.
A good local distance preserving method would fulfill $\norm{x_i - x_j}/\norm{x_k - x_l} \approx \norm{y_i - y_j}/\norm{y_k - y_l}$
or, equivalently, $\norm{x_i - x_j}^2/\norm{x_k - x_l}^2 \approx \norm{y_i - y_j}^2/\norm{y_k - y_l}^2$.
How does this quantity look for \tsne?
Let us assume that $\epsilon$ is small enough so that we can approximate $\exp(-x^2) \approx 1-x^2+O(x^4)$. We, thus, get
\begin{align}
  p_{ij} =& \frac{1}{2n} \left(\frac{1}{Z_i} \exp\left(-\frac{\norm{x_i - x_j}^2}{2\sigma^2_i}\right)\right. \\
  &~~~~+\left. \frac{1}{Z_j} \exp\left(-\frac{\norm{x_i - x_j}^2}{2\sigma^2_j}\right) \right)\\
  \approx& \frac{1}{2n}\left(\frac{1}{Z_i} + \frac{1}{Z_j}\right) \left(1- \left(\frac{1}{2\sigma^2_i} + \frac{1}{2\sigma^2_j} \right) \norm{x_i - x_j}^2 \right)\;,
\end{align}
where $Z_i :=  \sum_{k\neq i}\exp(-\norm{x_i - x_k}_2^2/(2\sigma_i^2)) $ denotes the corresponding normalization constant.
The same arguments also apply to the distance $\norm{x_k - x_l}$.
Similarly, for $q$ we can approximate 
\begin{align}
  q_{ij} & = \frac{1}{Z_q} \left( 1 + \norm{y_i - y_j}^{2} \right)^{-1} \approx \frac{1}{Z_q} \left( 1 - \norm{y_i - y_j}^{2} \right) \;,
\end{align}
with $Z_q = \sum_{m\neq p}(1+\norm{x_m - x_p}_2^2)^{-1}$.
Next, we will further employ the assumption that the \tsne optimization was successful so that $q_{ij} \approx p_{ij}$ and $q_{kl} \approx p_{kl}$.
In combination with our above approximations, this leads to the relation
\begin{align}
    \norm{x_i - x_j}^2 \approx 2 \widetilde{\sigma}^2_{ij} \left( (1-c_{ij}) + c_{ij}  \norm{y_i - y_j}^{2}\right)\; ,
\end{align}
with $c_{ij} := \frac{2n Z_i Z_j}{(Z_i + Z_j) Z_q}$ and $\widetilde{\sigma}^2_{ij} := \frac{\sigma^2_i\sigma^2_j}{\sigma^2_i+\sigma^2_j}$. 
This already highlights the general problem with \tsne: 
The constant and scaling factor of small distances depends on the neighbourhood of $i$ and $j$.
To make the problem more explicit let us study our quantity of interest, the preservation of relative distances:
\begin{align}
\frac{\norm{x_i - x_j}^2}{\norm{x_k - x_l}^2} 
& =  \frac{\widetilde{\sigma}^2_{ij} c_{ij}}{\widetilde{\sigma}^2_{kl} c_{kl}} \frac{  \frac{1}{c_{ij}} - 1 + \norm{y_i - y_j}^{2}}{ \frac{1}{c_{kl}} - 1 + \norm{y_k - y_l}^{2}} \; .
\end{align}
To simplify the expression note that since $\norm{x_i - x_j}$ is small, the local neighbourhood is similar, thus $\sigma_i \approx \sigma_j$ and therefore $Z_i \approx Z_j$. We get
\begin{align}
\frac{\norm{x_i - x_j}^2}{\norm{x_k - x_l}^2} 
& =  \frac{\sigma^2_i Z_i}{\sigma^2_k Z_k} \frac{  \frac{Z_q}{n Z_i} - 1 + \norm{y_i - y_j}^{2}}{ \frac{Z_q}{n Z_k} - 1 + \norm{y_k - y_l}^{2}} \; .
\end{align}
Intuitively, this means that unless the clusters are similar in density or size, i.e., $\sigma_i \approx \sigma_k$ and $Z_i \approx Z_k$, we can not preserve relative distances. Distances are scaled by a quantity that is inversely proportional to the high-dimensional variances.
A natural fix to these issues would therefore be to scale $\norm{y_i - y_j}^{2}$ proportional to this inverse of $\widetilde{\sigma}^2_{ij}$ (or any form of mean of $\sigma^2_i$ and $\sigma^2_j$
) and to choose $\sigma^2_i$ so that $c_{ij} \approx 1$ or at least $Z_i n \approx Z_q$.

\subsection{Preserving densities with \ourmethod}

As discussed above, to properly model relative densities, we need a distribution for our low-dimensional point pairs that properly reflect the density of their neighbourhood.
Conceptually, we want to map the distances of close neighbours of points in differently dense regions in $X$ to regions in $Y$ that show a similar relative difference in scale.

Based on the above insights, for a pair of points $i,j$ we define the  scaling factor for low-dimensional distances as
\begin{align}
 \gamma_{ij} &= \frac{\left((\sigma_i + \sigma_j)^2\right)^{-1}}{\max_{k,l}\left((\sigma_k + \sigma_l)^2\right)^{-1}} \; ,
\end{align}
which is a scaling factor that is inverse proportional to the squared average deviation of $i,j$, $1/(\sigma_i+\sigma_j)^2$, and normalized to have maximum value of $1$.

By incorporating the scaling into low-dimensional probabilities $q_{ij}$, we enable learning of relative densities as
\begin{align*}
q_{ij} &= \frac{(1+\gamma_{ij}\norm{x_i - x_j}_2^2)^{-1}}{\sum_{k\neq l}(1+\gamma_{kl}\norm{x_k - x_l}_2^2)^{-1}} \;.
\end{align*}
We further adapt the high-dimensional probabilities to be defined analogous to our low-dimensional probabilities in terms of the symmetry of the scaling factor.
That is, we make the distribution $p_{j|i}$ and, hence, $p_{ij}$ dependent on the neighbourhood of both $i$ and $j$, by using $\sigma_{ij}^2 = \left(1/2(\sigma_i + \sigma_j)\right)^2$.
This not only makes the distributions more comparable, but also allows us to analyze the behaviour of this method theoretically.
We thus get
\[
p_{j|i} = \frac{\exp(-\norm{x_i - x_j}_2^2/(2\sigma_{ij}^2))}{\sum_{k\neq i}\exp(-\norm{x_i - x_k}_2^2/(2\sigma_{ik}^2))}\; ,
\]
and symmetrize the distributions as in vanilla \tsne 
\[
 p_{ij}=\frac{p_{j|i} + p_{i|j}}{2n}\;.
\]
Deriving the KL-divergence on this new probability distributions with respect to $Y$, we get
\[
\frac{\partial KL(P ~||~ Q)}{\partial y_i} = \frac{4\sum_j (p_{ij} - q_{ij})(y_i - y_j)\gamma_{ij}}{(1+\gamma_{ij}\norm{y_i - y_j}_2^2)}\;.
\]
We give the derivation in App.~\ref{app:derivative}.
Based on this gradient, we can optimize for an embedding $Y$ by gradient descent. 
We call this method \ourmethod for density preserving \tsne, and give pseudocode in Alg.~\ref{alg:ours}.
By design, it closely resembles \tsne and comes with the same computational costs of $O(n^2T)$ for data of $n$ samples and descent for $T$ iterations.
In practice, we can make use of established ways to speed up and improve the optimization, such as early exaggeration and PCA initialization, both improving formation of natural clusters of the data in the embedding, and hence speeding up the overall computations. We refer to \citet{maaten2008visualizing} and \citet{kobak2019art} for details.

\renewcommand{\algorithmiccomment}[1]{\hfill// #1}
\renewcommand{\algorithmicrequire}[1]{\textbf{Input:} #1}
\renewcommand{\algorithmicensure}[1]{\textbf{Output:} #1}
\begin{algorithm}
   \caption{\ourmethod}\label{alg:ours}
\begin{algorithmic}[1]
    \REQUIRE{Data $X$, perplexity $k$, iterations $T$, learning rate $\mu$, momentum $\delta$}
    \ENSURE{Embedding $Y$}
    \STATE compute $P$ \COMMENT{Use symmetrized $\sigma_{ij}$}
    \STATE compute $\gamma_{ij}$ \COMMENT{Scaling factor $\gamma_{ij}$}
    \STATE $Y^{(0)} \leftarrow PCA(X,2)$ \COMMENT{Initialization of embedding}
    \STATE $Y^{(0)} \leftarrow .0001\frac{Y^{(0)}}{std(Y^{(0)})} $ \COMMENT{ \cite{kobak:21:tsneumapsame}}
    \FOR{$t= 1\ldots T$}
        \STATE compute $Q$ \COMMENT{Use scaling $\gamma_{ij}$}
        \STATE compute $\frac{\partial KL(P ~||~ Q)}{\partial Y}$
        \STATE $Y^{(t)} \leftarrow Y^{(t-1)} + \gamma\frac{\partial KL(P ~||~ Q)}{\partial Y} + \delta(Y^{(t-1)} - Y^{(t-2)})$
    \ENDFOR
    \STATE \textbf{return} $Y^{(T)}$
\end{algorithmic}
\end{algorithm}

\subsubsection*{Theoretical density preservation}

\ourmethod is able to address the outlined issues of preserving relative densities approximately just by rescaling the distances in $q$ with a variance $\gamma_{ij}$ that is proportional to $\sigma^2_{ij}$.
Recall that $p_{ij} = w_{ij} \exp\left(-\norm{x_i - x_j}^2/(2\sigma^2_{ij})\right)$, where $w_{ij} = \frac{1}{Z_p} \left(\frac{1}{Z_i} + \frac{1}{Z_j} \right) $ with a modified definition of $Z_i := \sum_{k\neq i}\exp(-\norm{x_i - x_k}_2^2/(2\sigma_{ik}^2))$ and a global normalization constant $Z_p$. 
Furthermore,  $q_{ij} = \frac{\left( 1 + \gamma_{ij}\norm{y_i - y_j}^{2} \right)^{-1}}{Z_q} $, with $Z_q = \sum_{k\neq l}(1+\gamma_{kl}\norm{x_k - x_l}_2^2)^{-1}$.

Considering close points $i,j$ and that
$p_{ij} \approx q_{ij}$, we can solve $p_{ij}$ for the distance
\begin{align}
  &\norm{x_i - x_j}^2 \\
  & \approx 2 \sigma^2_{ij} \left( \log\left(w_{ij} Z_q \right) + \log\left(1 + \gamma_{ij}\norm{y_i - y_j}^{2}\right)\right) \\
  & = \frac{2 \sigma^2_{ij}}{\gamma^{-1}_{ij}} \left( \log\left(w_{ij} Z_q \right) \gamma^{-1}_{ij} +\gamma^{-1}_{ij} \log\left(1 + \gamma_{ij}\norm{y_i - y_j}^{2}\right)\right)\\
  & \approx  \frac{2 \sigma^2_{ij}}{\gamma^{-1}_{ij}} \left( \log\left(w_{ij} Z_q \right) \gamma^{-1}_{ij} + \norm{y_i - y_j}^{2}\right),
\end{align}
where the last approximation holds for sufficiently small $\gamma_{ij}\norm{y_i - y_j}^{2}$.
\ourmethod sets $\gamma_{ij} = \lambda( \sigma^2_{ij})^{-1}$ for a $\lambda > 0$ so that we receive
\begin{align}
  &\norm{x_i - x_j}^2\\
  & \approx \frac{2 \sigma^2_{ij}}{\gamma^{-1}_{ij}} \left( \log\left(w_{ij} Z_q \right) \gamma^{-1}_{ij} +  \gamma^{-1}_{ij} \log\left(1 + \gamma_{ij}\norm{y_i - y_j}^{2}\right)\right)\\
  & \approx 2\lambda \Big( \log\left(w_{ij} Z_q \right) \lambda^{-1} \sigma^2_{ij} \\
  &\qquad\quad+  \lambda^{-1} \sigma^2_{ij} \log\left(1 + \lambda( \sigma^2_{ij})^{-1}\norm{y_i - y_j}^{2}\right)\Big)\\
   & \approx  2\lambda \left( \log\left(w_{ij} Z_q \right) \lambda^{-1} \sigma^2_{ij} + \norm{y_i - y_j}^{2}\right),
\end{align}
where the last approximation applies solely to small distances. 
Note that the choice of $\lambda$ does not affect the embedding $y_p$ or the normalization constant $Z_q$, as the optimization could just return $\lambda y_p$ instead of $y_p$, thus yielding the same probability distribution $q$ with the same $Z_q$ for any $\lambda > 0$.
$\lambda$ also does not influence the scaling factor of relative local distances, as it cancels out:
\begin{align}
\frac{\norm{x_i - x_j}^2}{\norm{x_k - x_l}^2} 
& \approx  \frac{  \log\left(w_{ij} Z_q \right)  \lambda^{-1}\sigma^2_{ij}  + \norm{y_i - y_j}^{2}}{ \log\left(w_{kl} Z_q \right)  \lambda^{-1}\sigma^2_{kl} \lambda  + \norm{y_k - y_l}^{2}}\; . 
\end{align}
We are thus free to choose $\lambda$ such that the contribution of $\log\left(w_{ij} Z_q \right) \lambda^{-1} \sigma^2_{ij}$ or $\log\left(w_{kl} Z_q \right)  \lambda^{-1}\sigma^2_{kl}$ becomes irrelevant in comparison with $\norm{y_i - y_j}^{2}$ or $\norm{y_k - y_l}^{2}$. 
We conclude that for small enough $\lambda$, \ourmethod succeeds in preserving relative local distances, as 
$\frac{\norm{x_i - x_j}^2}{\norm{x_k - x_l}^2}  \approx  \frac{ \norm{y_i - y_j}^{2}}{\norm{y_k - y_l}^{2}}$
holds for any pair of small distances $\norm{x_i - x_j}$ and $\norm{x_k - x_l}$.
Next, we show that \ourmethod also empirically preserves relative local distances.




\section{Experiments}

\input{figs/res_barplots.tex}

To evaluate \dtsne, we compare on both synthetic as well as real world data against the state-of-the-art in unsupervised low-dimensional embedding approaches \umap~\citep{mcinnes2018umap}, \tsne~\citep{maaten2008visualizing}, \lle~\citep{roweis2000nonlinear}, \largevis~\cite{tang:16:largevis}, and \ncvis~\cite{artamenkov:20:ncvis}. 
We consider benchmarks of gaussian and uniform mixtures, the (vectorized) MNIST dataset of handwritten digits~\citep{mnist}, two biological single-cell datasets~\citep{wong2016atlastcell, samusik2016phenotypesc}, and neural sentence embeddings of Amazon reviews~\citep{amazonreview}.

For \ourmethod and \tsne, we set the learning rate as $\mu=n/12$~\cite{belkina:19:tsnelr}, the momentum to $\delta=.5$ in the first $20$ iterations and to $\delta=.8$ afterwards~\cite{maaten2008visualizing}, and set the perplexity to $k=100$ in all experiments, which showed consistently good performance across all data.
For all other methods, we use the recommended parameter settings from the respective original publications.
Before embedding a given dataset, we project it to its first 50 principal components, a common practice to improve low-dimensional embeddings.
We use the OpenTSNE\footnote{\url{https://opentsne.readthedocs.io/en/latest/}} implementation for \tsne embeddings and use the original publicly available implementations for other methods. An implementation of \ourmethod and benchmark data generation is publicly available.\!\footnote{\oururl}
On all datasets, all methods take less than an hour to finish, with \umap, \tsne, \largevis, and \ncvis taking seconds to a few minutes and \ourmethod showing a slightly slower computation with 10-30 minutes depending on the data, which is due to a less-optimized code.
In particular, we stick to standard learning rates, use no fine-tuned learning rate schedule, or early stopping, leaving this in combination with other methods, such as FFT-based acceleration, for future work.

We compare all methods based on Pearson correlation $\rho$ between high- and low-dimensional distances, a common measure of embedding quality. Any correlation measures across all distances, however, places more importance on reconstruction of the global arrangement of data, and much less on the reconstruction of local structures, that we are usually interested in.
To evaluate how well such short-range (local) distances are preserved we compute the Pearson correlation $\rho_{knn}$ of distances of each point to its $k=100$ closest neighbours between high- and low-dimensional space.
Not only short-range distances but also the relative sizes and densities between different structures capture crucial information. We, hence, also evaluate how good relative densities are reconstructed in the embedding. For that we take for each point $i$ the radius $r_i$ of the (smallest) ball enclosing its $100$ neighbours, i.e., the distance to its $100^{th}$ neighbour. 
The ball serves as a proxy of how far neighbouring points are spread out in the space. 
As we are interested in reconstructing relative densities, we then take fractions of each pair of radii, $\frac{r_i}{r_j}$, and
measure the Pearson correlation $\rho_r$ between them in high- and low-dimensional space.

\subsection{Embeddings on synthetic benchmark data}

\input{figs/gaussian10d.tex}

We first benchmark all methods on synthetic data with known cluster structure, providing more details in the App.~\ref{app:data}.
Overall, we generate $6$ datasets varying different properties, such as number of samples per cluster, cluster variances, and generating distributions.
We first generate two 2D datasets with 3 Gaussian clusters each, one where each Gaussian has a different variance but we keep the number of samples per cluster fixed, and one where each Gaussian has same variance but the number of samples per cluster is different. 
These two datasets serve as the basis for Fig~\ref{fig:motivation_density} and Fig~\ref{fig:motivation_samples}, as they visually show the underlying problem of current low-dimensional embedding algorithms.

To evaluate all methods quantitatively, we generate more complex datasets in a 50 dimensional space. 
The first dataset is made of 3 Gaussian clusters with same scale but varying number of samples in each (G3-s). 
The second dataset has 3 Gaussian clusters with differing scale but same number of samples per cluster (G3-d). 
The third dataset contains 10 clusters with a different scale (G10-d) and the last dataset has 150 dimensions in which we place 5 clusters, each drawn from a Uniform distribution with a different scale (U5-d). 
The results are visualized in Fig.~\ref{fig:real_res}a-d.

When it comes to global reconstruction, \ourmethod performs on par with the best other methods.
More interestingly, when looking at quality of local reconstruction $\rho_{knn}$, which tells how well the actual local structures that we are interested in are preserved, we get a different picture. \lle still performs worst, yet all other competitors also show comparably bad performance, for example achieving only single-figure correlations on G3-d.
Only \ourmethod achieves consistently high quality when it comes to local reconstruction.
This trend is even more extreme when looking at preservation of relative densities $\rho_r$, showing that all except \ourmethod fail to recover densities.
While these datasets were challenging and also \ourmethod does not achieve perfect reconstruction, for example for G3-s, the state-of-the-art does not maintain any difference in cluster size at all.
This becomes also evident when looking at the visualization of the embeddings, for example for G10-d given in Fig.~\ref{fig:synthG10}.

\subsection{Embeddings of real world data}

Among our real world datasets, Amazon reviews is the most challenging to embed, as the reviews contain colloquial language and abbreviations and have frequent grammatical or spelling mistakes.
This challenging problem also reflects in the performances, regardless of tool or metric (see Fig.~\ref{fig:real_res}e).
Surprisingly, \tsne performs best. \ourmethod shows almost similar performance when it comes to reconstructing neigbourhoods and is the only other tool with decent performance when it comes to reconstructing relative densities.
Intriguingly, even though \tsne performs so well, it at the same time gives the least informative clusters -- only Luxury and Beauty products are clearly separated from the rest -- whereas other methods such as \ourmethod provide slightly better separation of individual clusters (see App.~Fig.~\ref{fig:amazon}). These clusters reveal informative sub-classes of products, such as knitting and crocheting, or shooting.

For MNIST embeddings, \ourmethod consistently performs best across all measures (see Fig.~\ref{fig:real_res}f). Except for \lle, which has overall bad performance, the other methods perform decently in terms of reconstructing global distances and neighbourhoods, with \tsne being the best of the competitors. Yet, consistent with our findings on synthetic data, we see that the state-of-the-art is not able to reconstruct relative densities well. Out of those competitors \tsne performs decently with $\rho_r=.38$, yet has a wide gap to the $.67$ achieved by \ourmethod. For the interested reader, we provide visualizations of the embeddings in App.~Fig.~\ref{fig:mnist}.

On Samusik and Wong data, we see a similar trend for global and neighbourhood reconstruction, with \ourmethod the best and competitors performing well, with a larger gap to \ourmethod in terms of neighbourhood reconstruction (see Fig.~\ref{fig:real_res}g,h).
Consistent with the literature~\cite{kobak:21:tsneumapsame}, we see that neither \umap nor \tsne is consistently better than the other.
When it comes to the reconstruction of relative densities, all methods fare better than on the MNIST data, yet have a substantially worse density reconstruction than \ourmethod. We provide a visualization of Samusik embeddings in App.~Fig.~\ref{fig:samusik}, where we observe that certain cell type clusters, such as pDCs and different T-cell types, are compacted in state-of-the-art embeddings, likely due to their small relative proportion in the overall data. In \ourmethod we see that those clusters are not that small in comparison to others once corrected for density -- these cell types likely have similar heterogeneity than others and are not as specialized as \umap or \tsne suggest.

\section{Discussion \& Conclusion}

We considered the problem of finding low-dimensional embeddings that capture the main regularities of high-dimensional data.
On a simple benchmark, we showed that the state-of-the-art methods fail to capture local densities at all and provided theoretical arguments on why this is the case.
Based on our findings, we proposed \ourmethod, a stochastic neighbourhood embedding approach that overcomes these issues by accounting for local variations in  data.

As opposed to the state-of-the-art, \ourmethod not only theoretically preserves relative density differences.
In extensive empirical experiments including synthetic benchmark as well as real world data, we also showed that \ourmethod faithfully reconstructs relative differences in local distributions, such as differently sized clusters. \ourmethod also quantitatively preserves local distances better than the state-of-the-art while yielding similar overall reconstruction performance.

Our approach easily scales to data of thousands of samples and is, thus, ready to be used in the applications of standard genomics or natural language processing datasets. For exceptionally large datasets, it would be an interesting avenue for future research to explore how we can combine \ourmethod with recent advances in improving runtime for low-dimensional embeddings, such as those based on FFTs~\cite{linderman2019fast}.

\ourmethod represents a first solution to low-dimensional embeddings that preserves relative local densities. We, hence, open up the analysis of low-dimensional embeddings and their visualizations with respect to cluster differences and densities. 
This could, for example, be used by experts as an indicator of heterogeneity or specialization of cell types, which are evident as clusters in embeddings of single-cell transcriptomics data.

\section*{Acknowledgements}
The authors thank Daniel Kindler for the insightful discussions and running preliminary experiments. 

\bibliography{lowdimembed.bib}
\bibliographystyle{icml2023}


\newpage
\appendix
\onecolumn
\section{\ourmethod gradient}
\label{app:derivative}

In the following, we derive the gradient for \ourmethod. 
In particular, we compute the derivative of the Kullback-Leibler ($KL$) divergence of the high and low dimensional probability distributions $P,Q$ with respect to the embedding $Y$.
For ease of notation, let
\begin{align*}
T_{ij}^{-1} &= (1+\gamma_{ij}\norm{x_i - x_j}_2^2)^{-1}\; ~and \\
Z_q &= \sum_{k\neq l}(1+\gamma_{kl}\norm{x_k - x_l}_2^2)^{-1} \;, ~then\\
q_{ij} &= \frac{T_{ij}^{-1}}{Z_q}\;
\end{align*}
and note that $T_{ij}^{-1}=T_{ji}^{-1}$.
Recall the loss function is given by the KL-divergence
\[
C = KL(P ~||~ Q) = \sum_i \sum_j p_{ij} \log\left(\frac{p_{ij}}{q_{ij}}\right)\;,
\]
where $p_{ij}$ is only determined by the high-dimensional data $X$ and thus fixed throughout optimization. Reorganizing the KL-divergence, we obtain
\[
C = \sum_i \sum_j p_{ij} \log\left(p_{ij}\right) - p_{ij}\log \left(q_{ij}\right)
= \sum_i \sum_j p_{ij} \log\left(p_{ij}\right) - p_{ij}\log \left(T_{ij}^{-1}\right) + p_{ij}\log \left(Z_q\right)\;.
\]
Deriving with respect to one sample $l$, we get
\begin{align*}
\frac{\partial C}{\partial y_l} &= \sum_i \sum_j \frac{\partial p_{ij} \log\left(p_{ij}\right)}{\partial y_l} - \frac{\partial p_{ij}\log \left(T_{ij}^{-1}\right)}{\partial y_l} + \frac{\partial p_{ij}\log \left(Z_q\right)}{\partial y_l} \\
&= \sum_i \sum_j - \frac{\partial p_{ij}\log \left(T_{ij}^{-1}\right)}{\partial y_l} + \frac{\partial p_{ij}\log \left(Z_q\right)}{\partial y_l}\;. \\
\end{align*}
In the following, we will analyze the left and right term separately. Starting with the left term, we can simplify by only looking at the terms of the sum dependent on $l$ and use basic rules of derivation
\begin{align*}
\sum_i \sum_j - \frac{\partial p_{ij}\log \left(T_{ij}^{-1}\right)}{\partial y_l} &= \sum_{k \neq l} - 2 p_{kl}\frac{\partial \log \left(T_{kl}^{-1}\right)}{\partial y_l} \\
&= \sum_{k \neq l} - 2 p_{kl} T_{kl} \frac{T_{kl}^{-1}}{\partial y_l}\\
&= \sum_{k \neq l} - 2 p_{kl} T_{kl}^{-1} (-2\gamma_{kl}(y_k - y_l))\\
&= 4\sum_{k \neq l} p_{kl} T_{kl}^{-1} \gamma_{kl}(y_k - y_l)\;.
\end{align*}
For the right term, we get
\begin{align*}
\sum_i \sum_j \frac{\partial p_{ij}\log \left(Z_q\right)}{\partial y_l} &= \sum_{k\neq l} Z_q^{-1}\frac{2 T_{kl}^{-1}}{\partial y_l}\\
&= \sum_{k\neq l} 2  Z_q^{-1} T_{kl}^{-2} (-2\gamma_{kl}(y_k - y_l)) \\
&= -4 \sum_{k\neq l} q_{kl} T_{kl}^{-1} \gamma_{kl}(y_k - y_l)\;,
\end{align*}
where we used that $\sum_{k\neq l} p_{kl} = 1$ the derivative of $Z_q$ can be split in the outer derivative $Z_q^{-1}$ and the inner derivative, for which only the two index combinations $kl$ and $lk$ are non-zero.

Combining the above, we arrive at 
\begin{align*}
\frac{\partial C}{\partial y_l} &= -4 \sum_{k \neq l} (p_{kl} - q_{kl}) T_{kl}^{-1} \gamma_{kl} (y_k - y_l)\;.
\end{align*}

\section{Experiments}
\label{app:data}
\input{figs/Uniform5d.tex}

In this section we provide additional information on the data generation and resulting embeddings. Numerical results are reported in Tab.~\ref{tab:results}. Additional visualizations for U5-d, which is the other more challenging synthetic benchmark dataset (i.e., where clusters can not be placed in a 2D plane) are given in Fig.~\ref{fig:synthU5}.
Visualizations of real world data are given further below.

\begin{table*}
\centering
\begin{tabular}{ll | cccc | cccc}
    \toprule
    
      && \multicolumn{4}{c}{Synthetic benchmark data} & \multicolumn{4}{c}{Real world data} \\
      Metric & Method &G3-s & G3-d & G10-d & U5-d & Amazon  & MNIST & Samusik & Wong \\
      \midrule
     \multirow{4}{*}{$\rho$} & \textbf{\ourmethod} &$ .99 $ & $ .95 $ & $ .62 $ & $ .79 $ & $ .27 $& $ .49 $ & $ .80 $ & $ .60 $ \\
     & \largevis &$ .98 $ & $ .95 $ & $ .59 $ & $ .85 $ & $ .35 $& $ .42 $ & $ .78 $ & $ .43 $ \\
     & \lle &$ .92 $ & $ .93 $ & $ .16 $ & $ .56 $ & $ .0 $&$ -.18 $ & $ .64 $ & $ .22 $ \\
     & \ncvis &$ .95 $ & $ .92 $ & $ .45 $ & $ .80 $ & $ .35 $&$ .36 $ & $ .63 $ & $ .45 $ \\
     & \tsne &$ .100 $ & $ .95 $ & $ .65 $ & $ .86 $ & $ .38 $&$ .48 $ & $ .77 $ & $ .60 $ \\
     & \umap &$ .98 $ & $ .95 $ & $ .57 $ & $ .85 $ & $ .33 $& $ .44 $ & $ .79 $ & $ .46 $ \\
     \midrule
     \multirow{4}{*}{$\rho_{knn}$} & \textbf{\ourmethod} &$ .74 $ & $ .81 $ & $ .71 $ & $ .82 $ & $ .54 $&$ .68 $ & $ .85 $ & $ .78 $ \\
     & \largevis &$ .56 $ & $ .08 $ & $ .39 $ & $ .13 $ & $ .52 $&$ .47 $ & $ .67 $ & $ .68 $ \\
     & \lle &$ .15 $ & $ -.08 $ & $ .21 $ & $ -.28 $ & $ -.28 $&$ -.40 $ & $ .64 $ & $ .52 $ \\
     & \ncvis &$ .59 $ & $ .09 $ & $ .40 $ & $ .10 $ & $ .44 $&$ .49 $ & $ .74 $ & $ .64 $ \\
     & \tsne &$ .70 $ & $ .12 $ & $ .40 $ & $ .12 $ & $ .62 $&$ .61 $ & $ .76 $ & $ .69 $ \\
     & \umap &$ .56 $ & $ .06 $ & $ .39 $ & $ .11 $ & $ .45 $&$ .44 $ & $ .65 $ & $ .66 $ \\
     \midrule
     \multirow{4}{*}{$\rho_r$} & \textbf{\ourmethod} &$ .31 $ & $ .88 $ & $ .91 $ & $ .89 $ & $ .22 $&$ .67 $ & $ .66 $ & $ .70 $ \\
     & \largevis &$ .02 $ & $ -.01 $ & $ -.01 $ & $ -.01 $ & $ .15 $&$ .08 $ & $ .31 $ & $ .42 $ \\
     & \lle &$ -.06 $ & $ -.02 $ & $ .08 $ & $ -.11 $ & $ -.16 $&$ -.24 $ & $ .34 $ & $ .22 $ \\
     & \ncvis &$ .16 $ & $ -.07 $ & $ -.08 $ & $ -.08 $ & $ .01 $&$ -.01 $ & $ .45 $ & $ .33 $ \\
     & \tsne &$ .20 $ & $ -.11 $ & $ .01 $ & $ -.09 $ & $ .34 $&$ .38 $ & $ .31 $ & $ .44 $ \\
     & \umap &$ -.02 $ & $ -.05 $ & $ -.02 $ & $ -.01 $ & $ .04 $&$ .02 $ & $ .40 $ & $ .38 $ \\
     \bottomrule
\end{tabular}
\caption{\textit{Results for synthetic and real-world data.} Synthetic data are generated as $k$ \textit{G}aussian or \textit{U}niform clusters and with varying \textit{s}ample-sizes or varying \textit{d}ensities across clusters. We report Spearman rank correlation between all high- and low-dimensional distances $\rho$ (global reconstruction), correlation between high- and low-dimensional distances of each point with its $100$ closest neighbours (local reconstruction), and correlation between radii of balls enclosing the $100$ neighbours of each point in high- and low-dimensional space (relative density reconstruction). }\label{tab:results}
\end{table*}


\subsection{Synthetic data}

We produced two different types of data, one where of the clusters are each distributed uniformly, and one where each cluster follows a Gaussian distribution. We varied the number of clusters $k$ the dimensionality of the data $d$, as well as the number of samples $s$ in each cluster or the spread of each cluster $d$.

\subsubsection*{2D data}
We generated two 2-dimensional dataset with 3 Gaussian clusters each. The Gaussian clusters had unit variance and were centered at $(10,0),(0,15),$ and $(-10,0)$, respectively.
For the first dataset, we drew $300$ points from each Gaussian and scaled the spread of the clusters by $1,2,4$ (i.e., multiply the centered data by this number), respectively.
For the second dataset we drew $100, 200, 500$, samples from the Gaussians, respectively, keeping the scale the same across clusters.

\subsubsection*{G3-s}
For this dataset we generated $3$ Gaussian clusters living in $50$ dimensions, each cluster distribution $c_i$ with mean drawn from $U(0,50)$ (each dimension iid from this uniform) and unit variance.
We then draw $200,400,600$ points from $c_1, c_2, c_3$, respectively, and scale the spread of the cluster by $2$ (i.e., multiply the centered data by 2).

\subsubsection*{G3-d}
For this dataset we generated $3$ Gaussian clusters living in $50$ dimensions, each cluster distribution $c_i$ with mean drawn from $U(0,50)$ and unit variance.
We then draw $300$ points from  each of the cluster distributions and scale the spread of the $c_1, c_2, c_3$ by $2, 4, 8$, respectively.

\subsubsection*{G10-d}
To look at data that is not easily projectable, i.e., the inter-cluster distances can be correctly modeled in 2D, for this dataset we generated $10$ Gaussian clusters living in $50$ dimensions, each cluster distribution $c_i$ again with mean drawn from $U(0,50)$ and unit variance.
We then draw $200$ points from each of the cluster distributions and scale the spread of the $c_1, \ldots, c_{10}$ by $1, \ldots, 10$, respectively.

\subsubsection*{U5-d}
To look at a different distribution and higher dimensional data, we generated $10$ Uniform clusters living in $150$ dimensions, each cluster distribution $c_i$ again with mean drawn from $U(0,50)$ and unit variance.
We then draw $200$ points from each of the cluster distributions and scale the spread of the $c_1, \ldots, c_{10}$ by $1, \ldots, 10$, respectively.

\subsection{Real data}

\begin{figure*}
\centering
\begin{subfigure}[t]{0.49\textwidth}
\centering
\includegraphics[width=9cm]{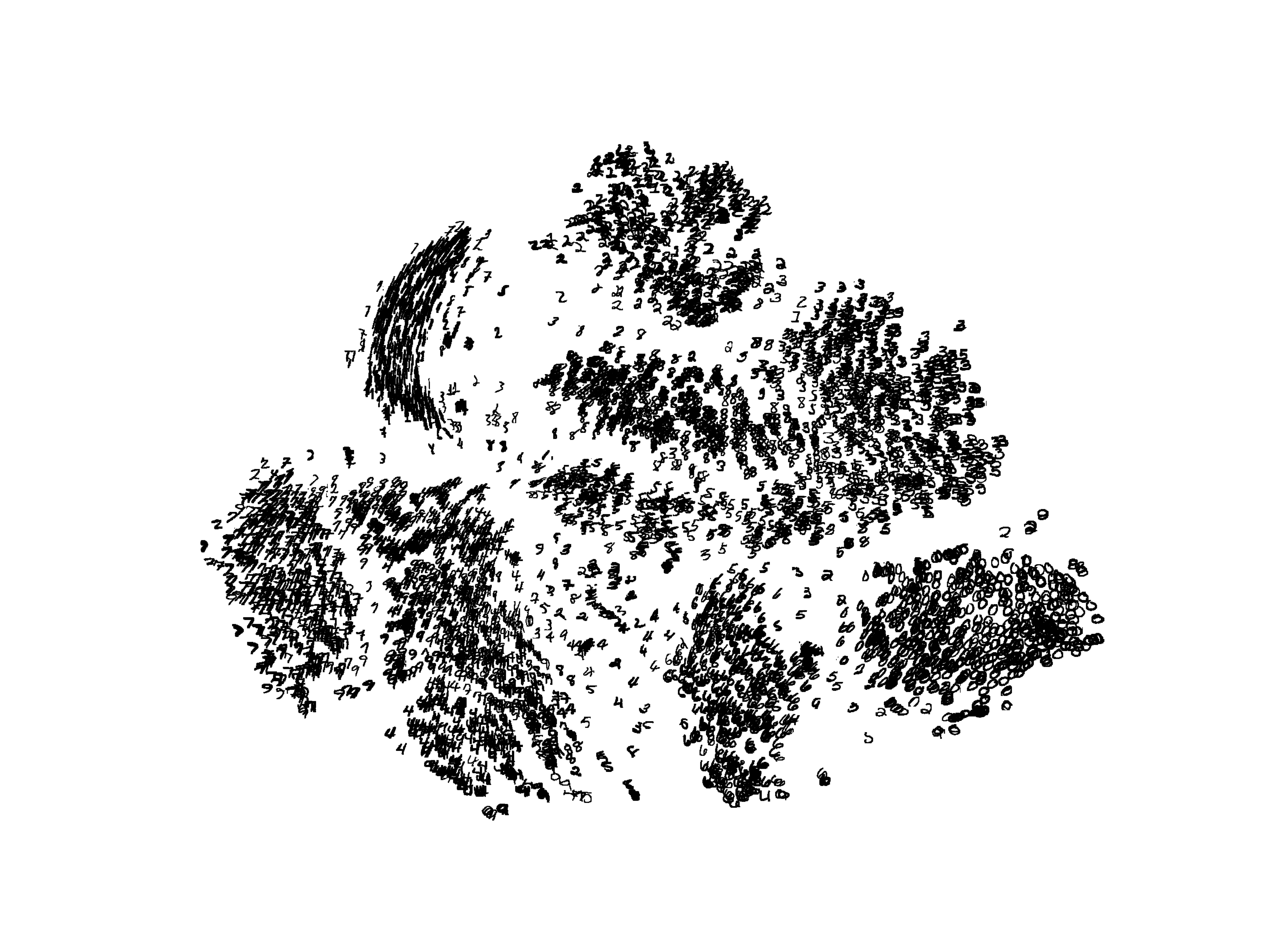}
\caption{\ourmethod}
\end{subfigure}
\begin{subfigure}[t]{0.49\textwidth}
\centering
\includegraphics[width=9cm]{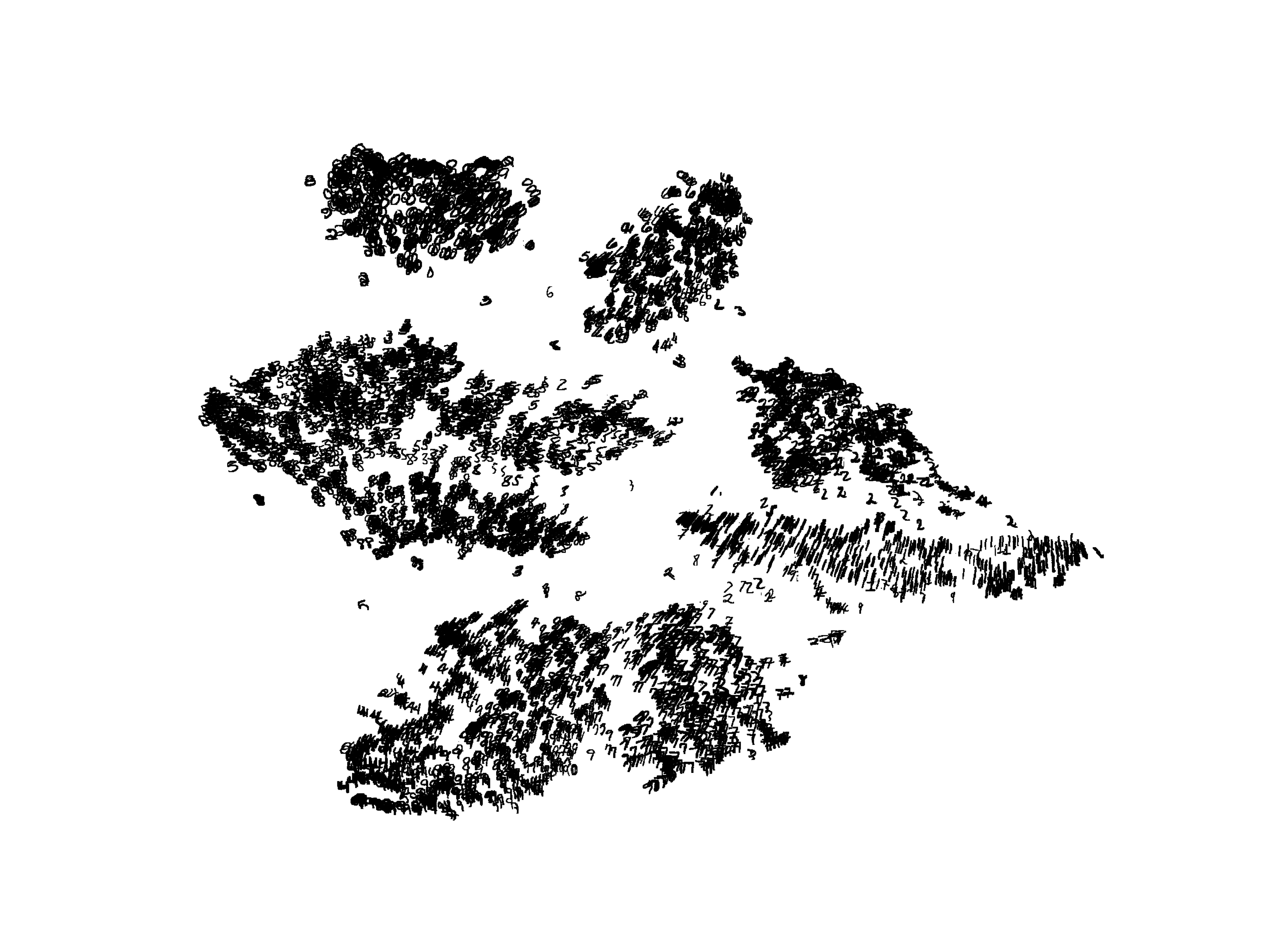}
\caption{\largevis}
\end{subfigure}
\begin{subfigure}[t]{0.49\textwidth}
\centering
\includegraphics[width=9cm]{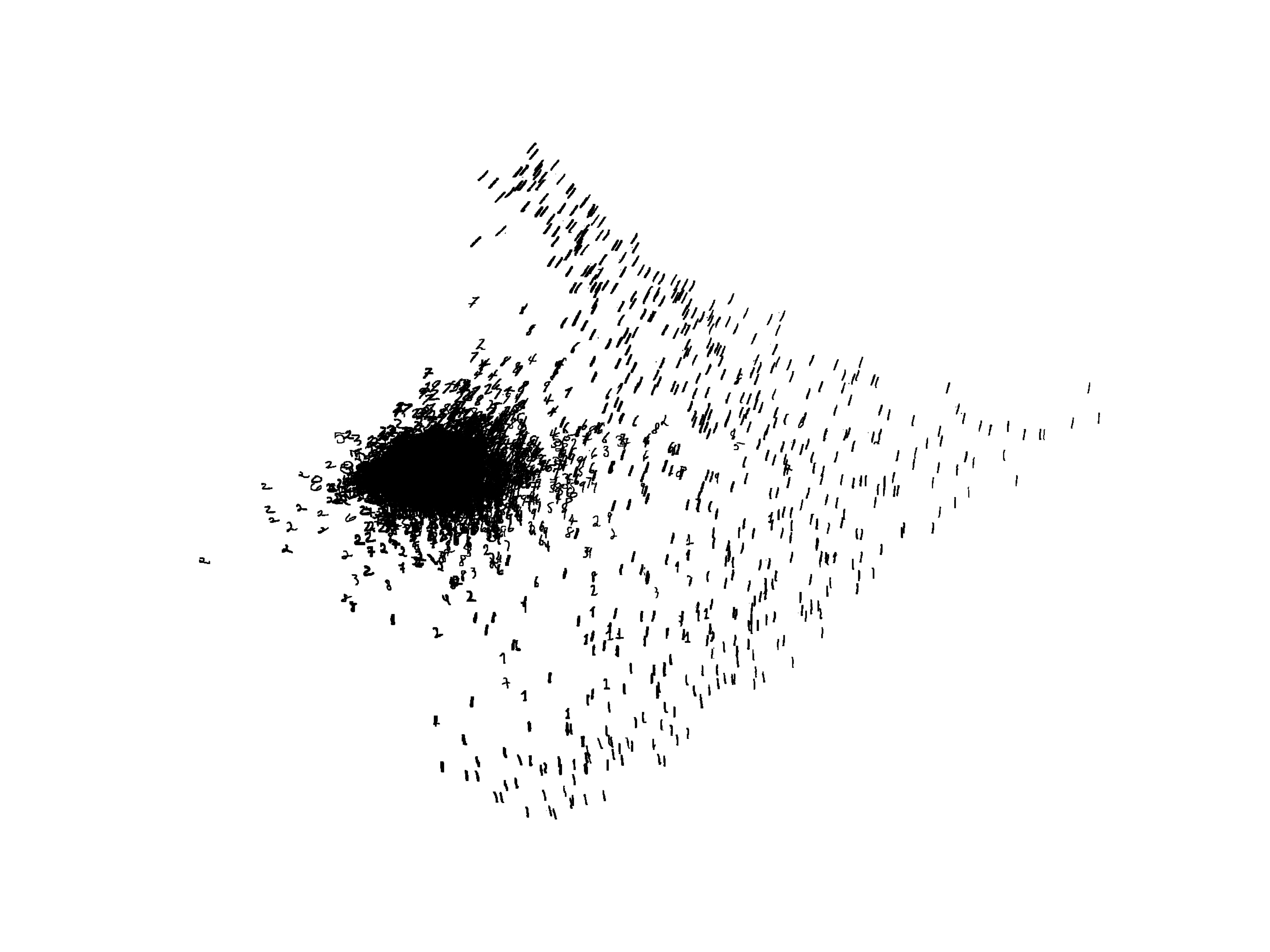}
\caption{\lle}
\end{subfigure}
\begin{subfigure}[t]{0.49\textwidth}
\centering
\includegraphics[width=9cm]{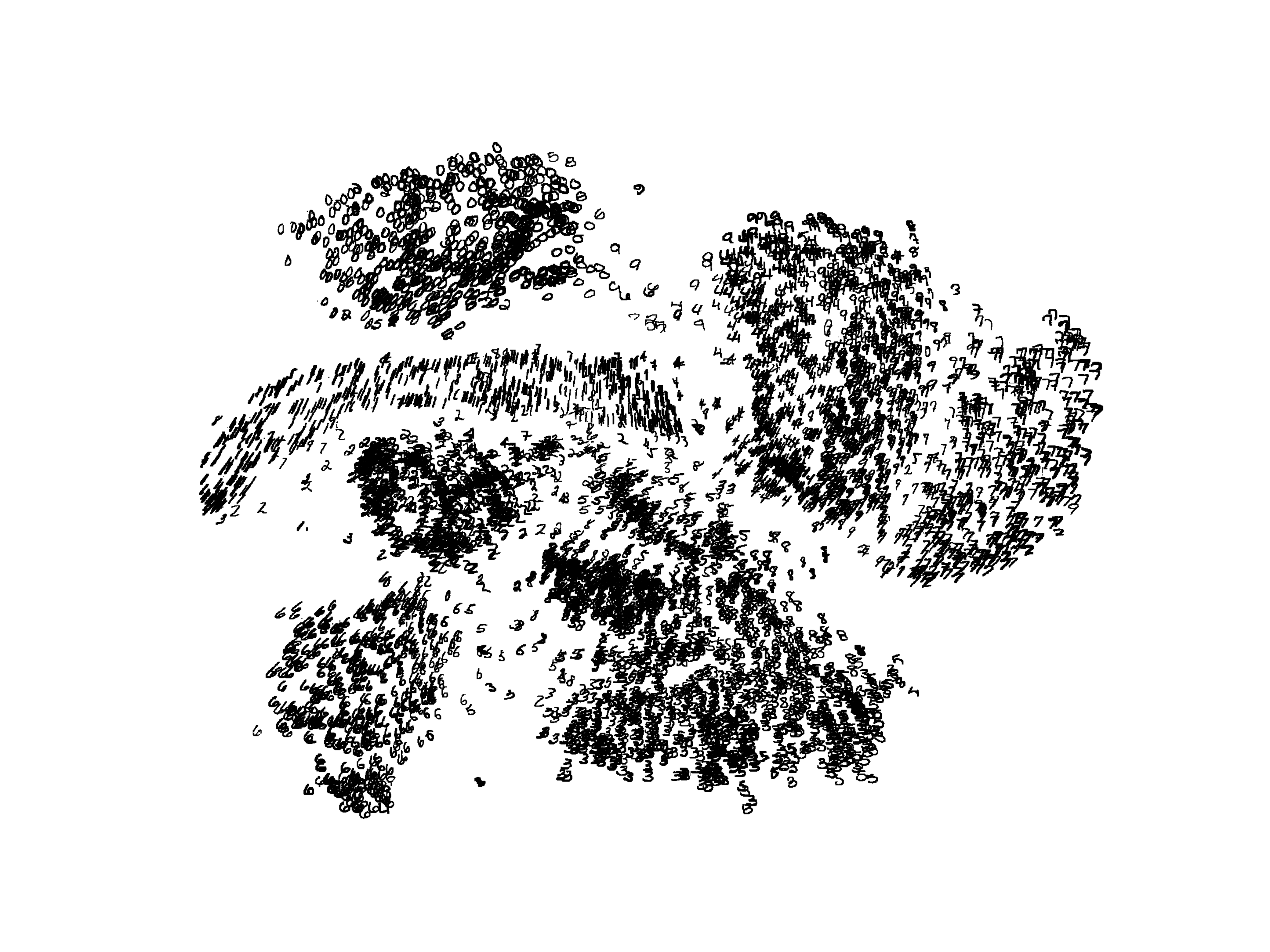}
\caption{\ncvis}
\end{subfigure}
\begin{subfigure}[t]{0.49\textwidth}
\centering
\includegraphics[width=9cm]{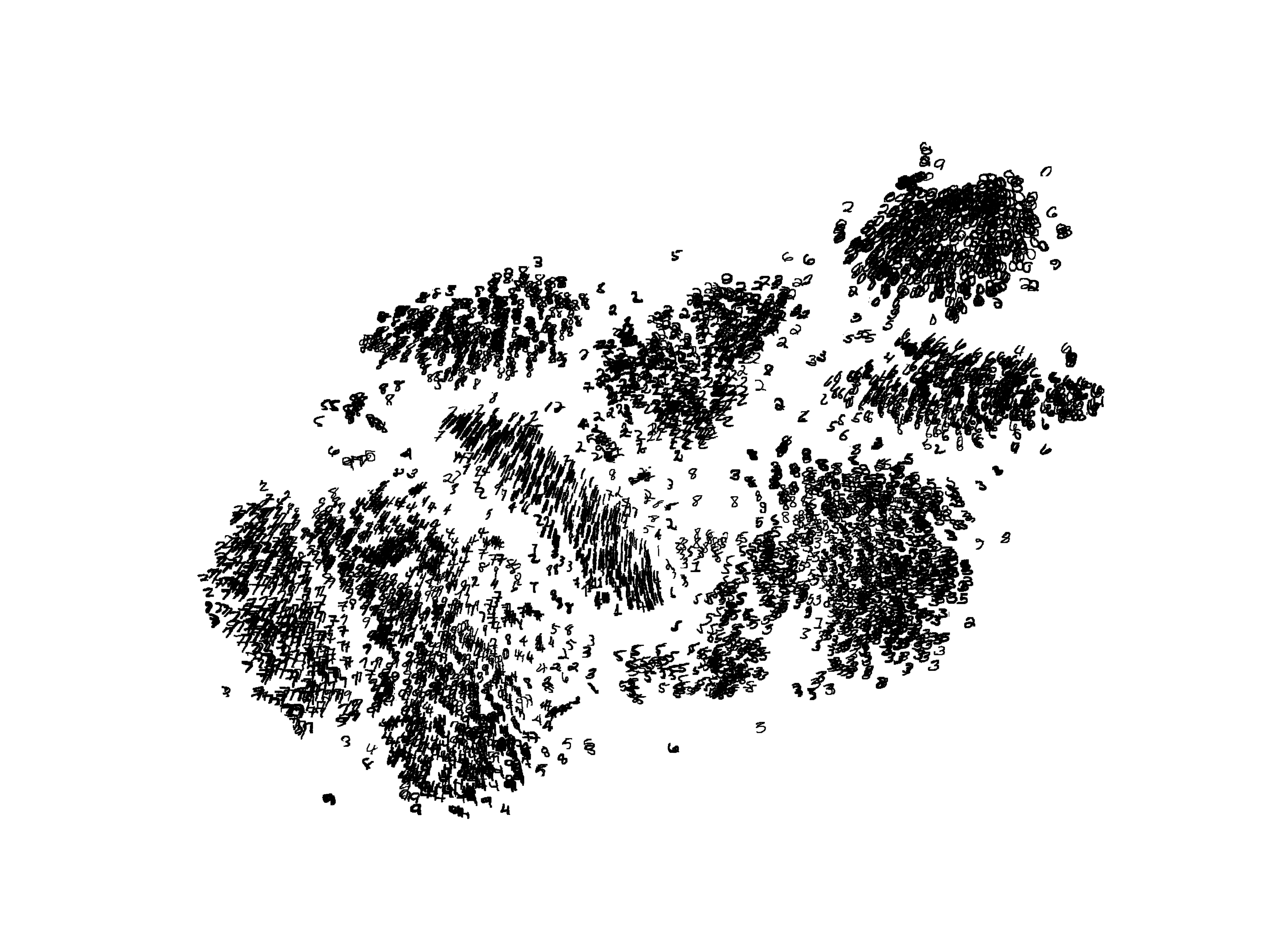}
\caption{\tsne}
\end{subfigure}
\begin{subfigure}[t]{0.49\textwidth}
\centering
\includegraphics[width=9cm]{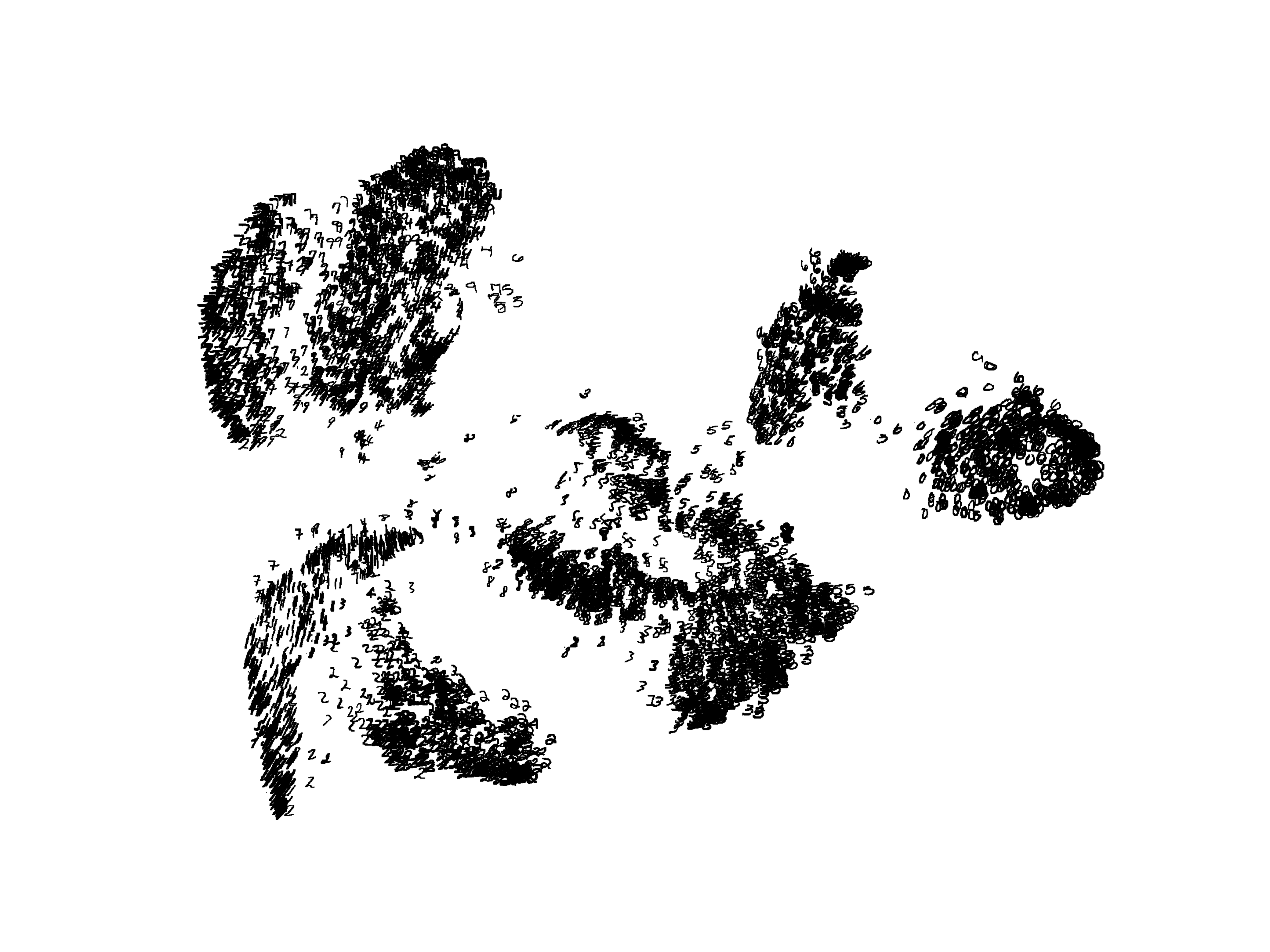}
\caption{\umap}
\end{subfigure}
\caption{\textit{MNIST embeddings.} Embedding of a random subset of 5000 samples from the MNIST dataset. Each sample is visualized as the original image stripped off its background, which allows to see inter-cluster dependencies such as curvatures and writing styles of digits.}\label{fig:mnist}
\end{figure*}
\begin{figure*}
\centering
\begin{subfigure}[t]{0.49\textwidth}
\centering
\includegraphics[width=6cm]{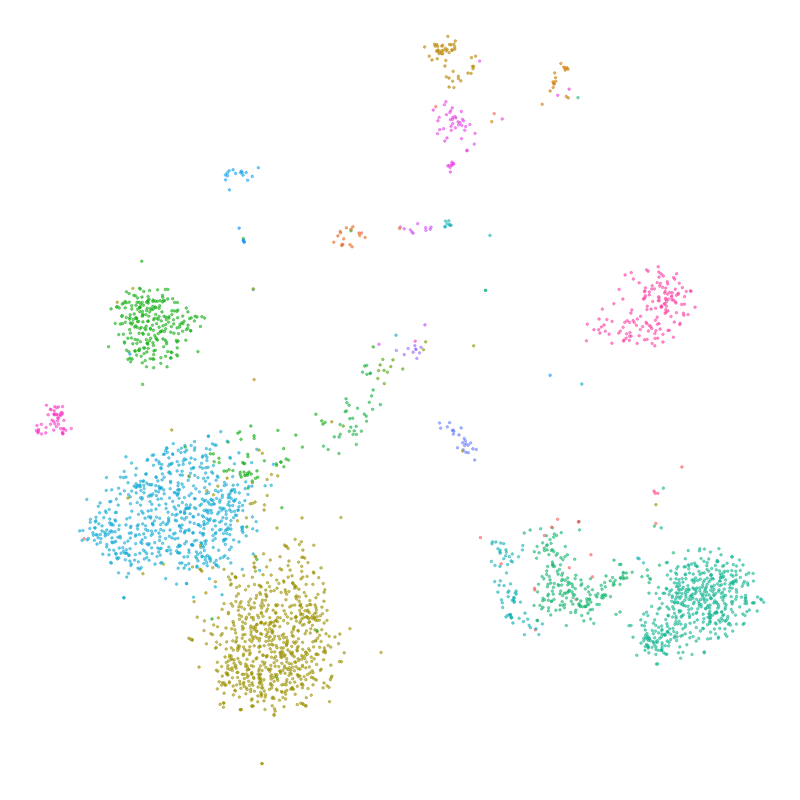}
\caption{\ourmethod}
\end{subfigure}
\begin{subfigure}[t]{0.49\textwidth}
\centering
\includegraphics[width=6cm]{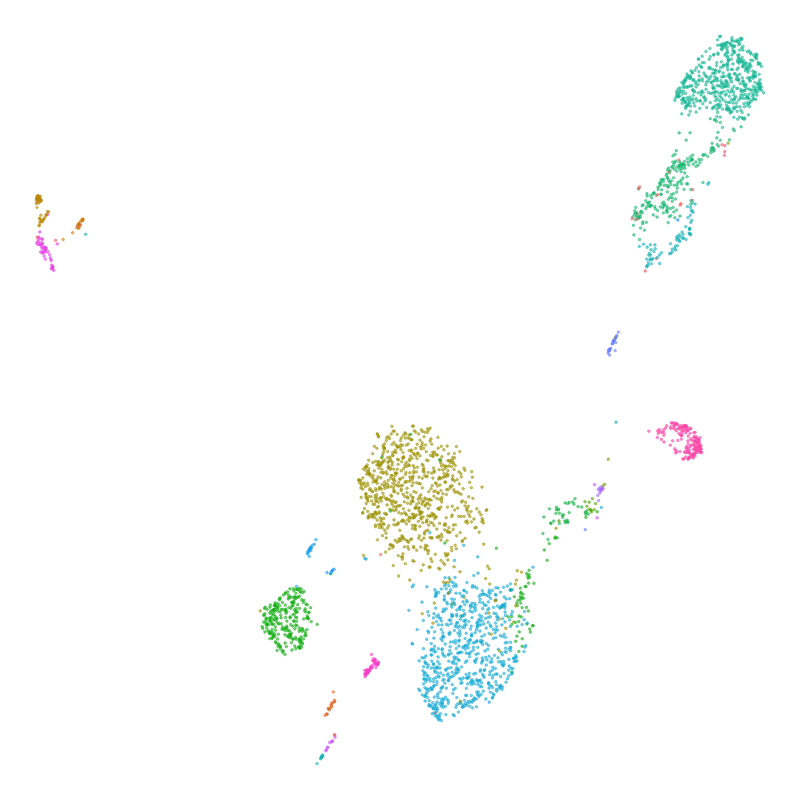}
\caption{\largevis}
\end{subfigure}
\begin{subfigure}[t]{0.49\textwidth}
\centering
\includegraphics[width=6cm]{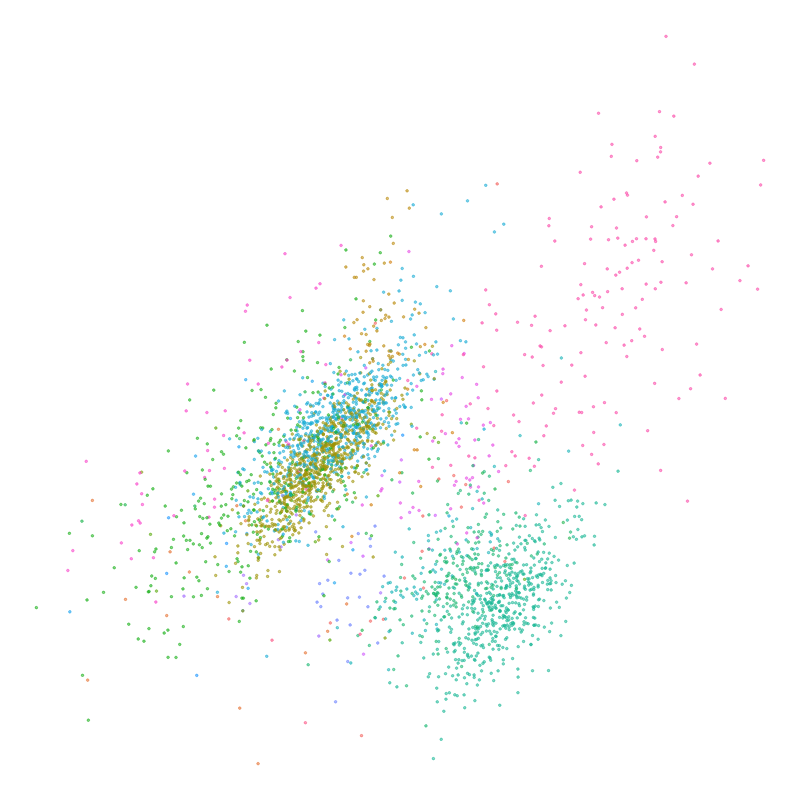}
\caption{\lle}
\end{subfigure}
\begin{subfigure}[t]{0.49\textwidth}
\centering
\includegraphics[width=6cm]{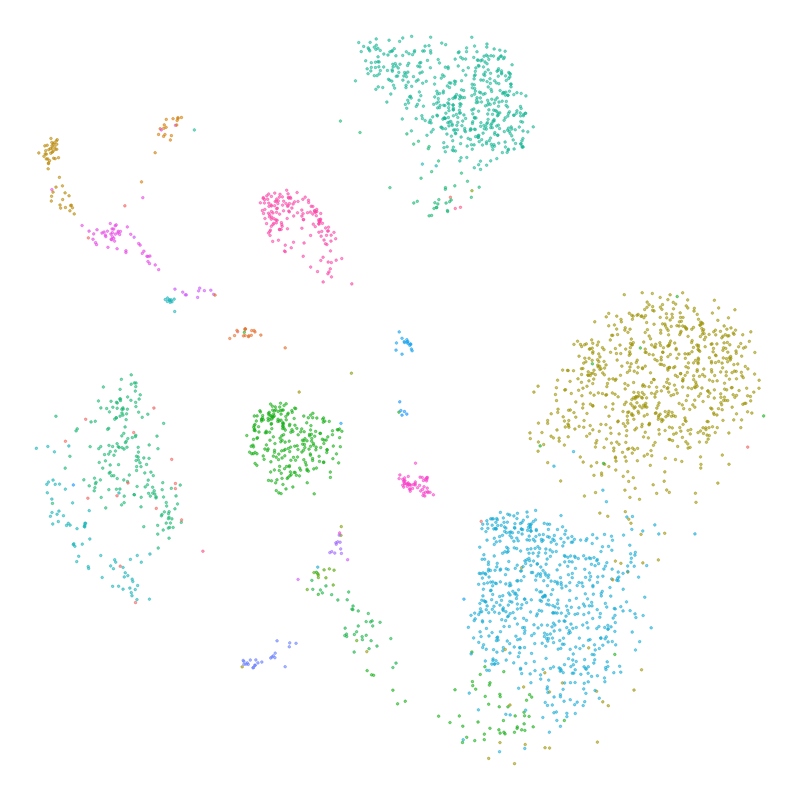}
\caption{\ncvis}
\end{subfigure}
\begin{subfigure}[t]{0.49\textwidth}
\centering
\includegraphics[width=6cm]{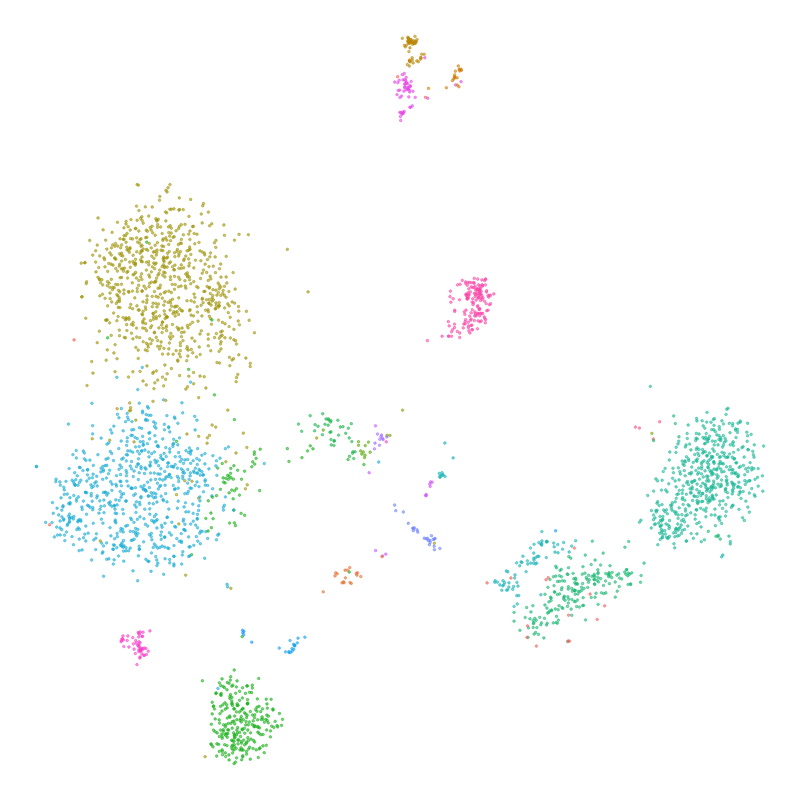}
\caption{\tsne}
\end{subfigure}
\begin{subfigure}[t]{0.49\textwidth}
\centering
\includegraphics[width=6cm]{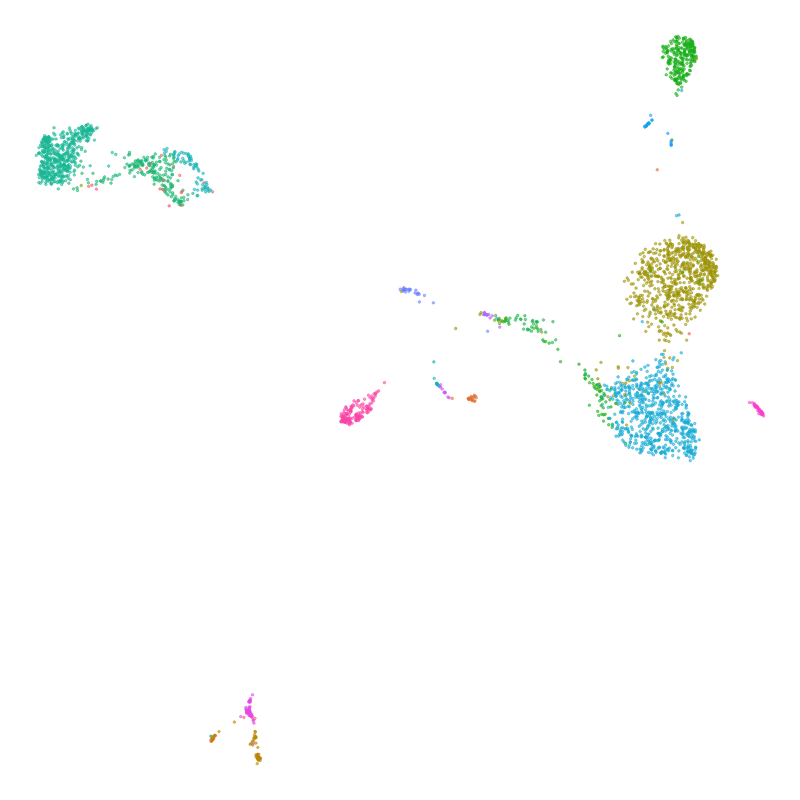}
\caption{\umap}
\end{subfigure}
\begin{subfigure}[t]{0.99\textwidth}
\centering
\includegraphics[width=16cm]{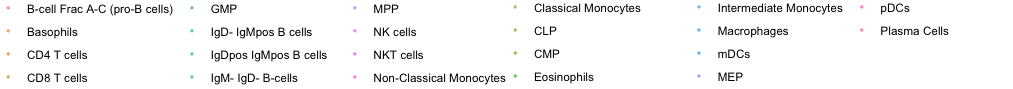}
\end{subfigure}
\caption{\textit{Samusik et al. single-cell data embeddings.} Embedding of a random subset of 5000 samples from the Samusik dataset. Samples are colored by cell type annotation from the original study. Cluster arrangement in all except \lle reflect hematopoiesis.}\label{fig:samusik}
\end{figure*}
\input{figs/amazon_reviews.tex}

For our comparison on real data, we considered $4$ datasets, a simple image benchmark, two biological single-cell datasets, and a neural sentence embedding of Amazon reviews.
For the image benchmark MNIST (Fig.~\ref{fig:mnist}) and both single-cell datasets we took a random subset of $5000$ samples from the original publications as referenced in the main manuscript, for MNIST we additionally vectorized the images. We provide the visualizations of embeddings for the Samusik et al. data in Fig.~\ref{fig:samusik}, Wong et al. did not have informative labels available for these embeddings.

For the Amazon Review dataset, we downloaded reviews for $8$ categories from \url{https://nijianmo.github.io/amazon/index.html} that are closely related: "Patio Lawn and Garden", "Tools and Home Improvement", "Industrial and Scientific", "Sports and Outdoors","Amazon fashion", "Arts and Crafts","Clothing, Shoes, Jewelry", and "Luxury Beauty".
We then sampled $5000$ reviews that had at least $15$ words in it, keeping original proportions of the categories intact. We set the threshold of $15$ words to keep only reviews that are more likely to be informative about a product, as there are many reviews that just read "Great product!!!" or "Can highly recommend!".
 We then use the Universal Sentence Encoder (\url{https://tfhub.dev/google/universal-sentence-encoder/4}) to obtain a 512-dimensional embedding of each review, resulting in the input data for our experiments.
 We give embeddings for the top $4$ methods only, to be able to fit on one page, in Fig.~\ref{fig:amazon}.

\end{document}